\documentclass[letterpaper, 10pt, conference]{ieeeconf}

\IEEEoverridecommandlockouts
\usepackage{amsmath,amssymb,amsthm}
\usepackage{mathtools}
\usepackage{bm}
\usepackage[ruled,vlined,linesnumbered]{algorithm2e}
\usepackage{cite}
\usepackage{hyperref}
\usepackage{booktabs}
\usepackage{graphicx}
\usepackage{caption}

\usepackage{dblfloatfix}

\DeclareMathOperator*{\argmin}{arg\,min}

\newcommand{\R}{\mathbb{R}}
\newcommand{\norm}[1]{\left\lVert #1 \right\rVert}

\newcommand{\proj}{\mathrm{proj}}
\DeclareMathOperator{\dist}{dist}

\title{\LARGE \bf
Contact-Implicit Stein Projected ADMM for Discovery of\\
Diverse Contact-Rich Manipulation Strategies
}

\author{Hrishikesh Sathyanarayan$^{\dagger}$, Christian Hughes$^{\dagger}$, and Ian Abraham$^{\dagger,\ddagger}$%
\thanks{$^\dagger$Department of Mechanical Engineering, Yale University, New Haven, CT, USA. {\tt\small \{hrishi.sathyanarayan, christian.hughes\}@yale.edu}}%
\thanks{$^\ddagger$Department of Electrical Engineering, University of Sydney, NSW, Australia.{\tt\small ian.abraham@sydney.edu.au}}%
}

\IEEEaftertitletext{%
\begin{center}
    \vspace{-2em}
    \includegraphics[width=\textwidth]{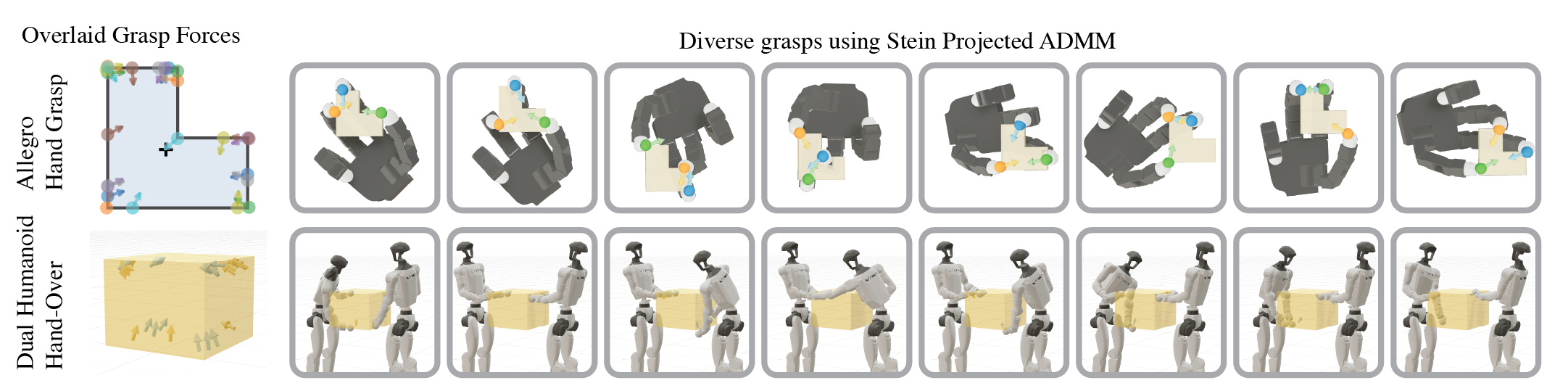}
    \captionof{figure}{
        This paper proposes Stein Projected ADMM, a method that discovers diverse, feasible contact-rich manipulation strategies by combining Stein variational inference and Alternating Direction Method of Multipliers (ADMM). Illustrated are diverse solutions for single-object grasps (top) and dual-humanoid stable hand-over (bottom).}
    \label{fig:teaser}
\end{center}
}

\begin{document}

\maketitle
\thispagestyle{empty}
\pagestyle{empty}

% \begin{figure*}
%     \centering
%     \includegraphics[width=\textwidth]{results/diverse_grasps_two_column.png}
%     \caption{
%         This paper proposes Stein Projected ADMM, a method that discovers diverse, feasible contact-rich manipulation strategies by combining Stein variational inference and Alternating Direction Method of Multipliers (ADMM). Illustrated are diverse solutions for single-object grasps (top) and dual-humanoid stable hand-over (bottom).
%     }
%     \label{fig:teaser}
%     \vspace{-0.2em}
% \end{figure*}

\begin{abstract}
    Contact-implicit trajectory optimization formulates contact-rich manipulation as a single constrained program; however, that single program run collapses onto one local optimum out of many equally valid contact modes, grasps, or push directions. 
    As a consequence, the resulting manipulation strategy is reluctant to change and sensitive to initialization.
    In order to promote robust manipulation, this paper investigates how contact-implicit solvers can discover diverse contact-rich strategies. 
    Our approach derives a variation of Consensus Alternating Direction Method of Multipliers (ADMM) combined with Stein variational inference methods to output a set of distinct contact-rich solutions. 
    We find that applying the Stein repulsive force to ADMM's split variable (rather than its primal form) allows for effective coverage over the set of feasible contact strategies without prematurely stalling the solver. 
    We demonstrate the effectiveness of our approach on a variety of contact-rich manipulation tasks, including pushing, grasping, and multi-robot handover. 
    Last, we find the proposed solver is simpler in form and capable of discovering unique contact modes when compared with existing solvers.
    Videos and code with examples are found in \url{https://anon-website-submission.github.io/stein-admm-website/}.
\end{abstract}

\section{Introduction}

    Contact-implicit trajectory optimization (CIO) has enabled robots to plan complex contact sequences directly without needing to pre-specify hybrid modes, building on earlier penalty- and continuation-based methods for hybrid mechanical trajectory optimization~\cite{Yunt2005SUMT, 10.1007/978-1-4020-6332-9_19}.
    CIO methods function by encoding contact/no-contact and stick/slip force transitions as decision variables subject to complementarity constraints that enforce physical realism (e.g., non-penetration, no force-at-a-distance, Coulomb friction) inside a nonlinear program.
    This unifies planning across contact modes, but it also means the feasible set is a union of many lower-dimensional manifolds (one per contact mode), each containing its own local optima.
    A single run of the solver terminates at one of these local optima, discarding other equally valid contact strategies (which face to push from, which fingers to grip with, which side to brace against).
    Consequently, the resulting manipulation strategies are sensitive to initialization, and reluctant to change to other equally valid contact strategies, even when the environment changes or the robot is perturbed.

    The underlying challenge is that these local minima are strong attractors; re-running the same solver from randomized initializations tends to collapse to the same local solution. 
    Even warm-starting from a different local optimum often fails to escape the basin of attraction.
    Thus, what is needed is a mechanism to encourage diversity in the solutions discovered by a single optimization formulation, rather than relying on multiple independent runs of the same solver.
    Thus, this paper proposes a new solver that combines the strengths of Stein Variational Inference (SVI) to promote diversity and recent CIO methods using the Alternating Direction Method of Multipliers (ADMM) to facilitate the discovery of diverse contact-rich manipulation strategies in a single run.

    % Stein Variational Inference (SVI) methods \cite{NIPS2016_b3ba8f1b} are a class of variational inference methods that approximate a target distribution.
    Stein Variational Inference (SVI) methods instantiate a deterministic gradient descent algorithm named Stein Variational Gradient Descent (SVGD) that maintains a population of particles and moves them to approximate a target distribution with a repulsive term that ensures particles remain distinct and diverse.
	This paper poses the target distribution as the Boltzmann distribution of the contact-implicit trajectory optimization Lagrangian.
	However, rather than encoding the constraints into SVGD as done in prior works, this paper uses a variable-splitting construction via ADMM that decouples the score term from the constraints and uses the repulsive term to encourage diversity within the feasible set in the particle population.
	As a result, we show that the proposed solver is able to discover multiple distinct contact-rich manipulation strategies on a number of tasks in a single run and avoids collapsing to a single local optimum.
	%% TODO: fix this sentence so it is not so targeted
	In addition, our solver is simpler in form than prior works, and does not require a kernel-space quadratic program or a merit-function line search to maintain feasibility.
	In summary, the contributions of this paper are:
	\begin{itemize}
		\item A novel Stein projected variable-splitting ADMM construction that injects diversity into ADMM's split variable rather than its primal variable;
        % so an exact per-outer-iteration projection guarantees feasibility independently of how strongly repulsion is tuned -- unlike prior constraint-aware SVGD methods, which fold feasibility and diversity into the same score.
		% \item A penalty-rescaling identity that keeps the scaled ADMM dual consistent under an adaptively growing penalty (Proposition~\ref{prop:rescale}), together with a structural failure mode of variable-splitting under max/min-structured (box) constraints that a direct, non-split penalty does not share.
		\item Demonstration of a distributed, contact-implicit trajectory solver whose simpler form facilitates discovery of diverse contact modes at lower computational cost; and
        \item Validation across four contact-rich manipulation tasks  (pushing, grasping, bimanual hand-over, and dynamic two-finger box pivoting).
	\end{itemize}
	% --- previous version, kept for comparison ---
	% \begin{itemize}
	% 	\item A new solver that combines Stein Variational Inference with ADMM to discover diverse contact-rich manipulation strategies in a single run.
	% 	\item A variable-splitting construction that decouples the Stein variational gradient from the constraints, allowing the repulsive term to encourage diversity within the feasible set.
	% 	\item Demonstration of the effectiveness of the proposed solver on a variety of contact-rich manipulation tasks, including pushing, grasping, and multi-robot handover.
	% \end{itemize}

	% The overall paper is organized as follows. Sections~\ref{sec:related_work} and~\ref{sec:prelims} give the necessary background and Section~\ref{sec:method} derives the proposed method; 
    % Section~\ref{sec:results} discusses empirical results and Section~\ref{sec:conclusion} provides a discussion of future directions.

\section{Related Work}
\label{sec:related_work}

	\subsection{Contact-Implicit Trajectory Optimization}

        Contact-implicit trajectory optimization (CIO) lifts contact forces as decision variables and enforces rigid-body contact through linear complementarity constraints~\cite{doi:10.1177/0278364913506757, doi:10.1177/0278364919849235}.
        This lets a nonlinear program reason jointly over poses, velocities, and contact forces, without requiring a pre-specified contact sequence.
        However, because feasible contact modes lie on distinct, lower-dimensional manifolds, CIO is highly sensitive to initialization and prone to collapsing onto a single local optimum.
        Existing solutions address this indirectly through Monte-Carlo restarts that diversify the initialization~\cite{toussaint2024nlpsamplingcombiningmcmc, doi:10.1177/02783649241273645}, or generative models that diversify the solutions directly~\cite{pmlr-v164-ortiz-haro22a}, which comes at the cost of extra training data and poor generalization to unseen constraints.
        Complementarity-free reformulations of contact dynamics~\cite{jin2025complementarityfreemulticontactmodelingoptimization} sidestep this multi-modal manifold structure entirely, but forgo the LCP's unified treatment of contact modes.
        What is missing is a mechanism that shares information across many CIO solves so that diversity emerges within the solver rather than from external sources.

	\subsection{Stein Variational Inference}

        Stein Variational Gradient Descent (SVGD) provides such a mechanism to encourage diversity within an optimization problem. 
        SVGD is derived from Stein variational inference where an ensemble of particles is evolved in parallel following the score of a target density while a kernel repulsion term prevents collapse onto a single mode~\cite{NIPS2016_b3ba8f1b, NEURIPS2018_fdaa09fc}.
        Extensions to high-dimensional settings project the Stein update onto a low-dimensional informed subspace~\cite{NEURIPS2020_14faf969}.
        However, SVGD assumes an unconstrained domain, and extending it to constrained domains is an open area of research.

        Recent advances integrate constraints into SVGD via a projection-based kernel that keeps particles tangent to a single equality manifold~\cite{NEURIPS2022_f092c842}, building on classical constraint-manifold motion planning~\cite{Berenson-2009-10209}, but does not extend to nonlinear inequalities or non-convex feasible sets.
        Other approaches use penalty-based methods that absorb constraints directly into the target density score through several mechanisms~\cite{10598358, tabor2025constrainedsteinvariationalgradient, li2026globalizedconstrainedsteinvariational}.
        All of these couple the target density and diversity within the same update, altering the particle flow in ways that can compromise optimality and feasibility.
        Beyond constrained sampling, SVI-based particle methods have also been applied to robot exploration and control under uncertainty~\cite{Lee-RSS-24, NEURIPS2025_6b4067db, SathyanarayanH-RSS-26, Sathyanarayan2026Stein}.

        This paper integrates diversity through ADMM where the repulsive gradient of SVGD enters as a desired projection for the split variable, encouraging diversity while the primal step maintains feasibility, reaching consensus only once the split variable aligns with the primal solution. 
        The underlying mechanism for encouraging diversity is then simpler, enabling variability within the solver without compromising feasibility and optimality.

\section{Preliminaries}
\label{sec:prelims}

    \subsection{Contact Dynamics via Complementarity}
    \label{sec:manipulator}

        The dynamics of an articulated rigid body is governed by the manipulator equation written in generalized discrete-time coordinates $q_k \in \R^{n_q}$ with velocity $v_k = \dot q_k \in \R^{n_v}$ and control $u_k \in \R^{n_u}$ as 
        \begin{align}
            M(q_k)\,\left(v_{k+1} - v_k\right) &+ b(q_k,v_k) =
            B u_k + \sum_{i} J_{c,i}(q_k)^\top \bar \lambda_{i,k}, \nonumber \\ 
            &q_{k+1} = q_k \oplus  \, v_{k+1},
        \label{eq:manipulator}
        \end{align}
        where $\oplus$ denotes the configuration update operator\footnote{This is a general addition to account for rigid body updates in SE(3).}, $\bar \lambda_{i,k} = [\lambda_{n,i,k};\lambda_{t,i,k}]
        \in \R^{1+n_t}$ is the contact impulse at time step $k$ that consists of a non-negative normal force $\lambda_{n,i,k} \in
        \R$ and a friction force $\lambda_{t,i,k} \in \R^{n_t}$ subject to the normal complementarity
        condition $\lambda_{n,i,k} \, \phi_i(q_k) = 0$, $\lambda_{n,i,k} \geq 0$, $\phi_i(q_k) \geq 0$, and Coulomb friction cone which is compactly expressed as
        \begin{equation}
            0 \leq \lambda_{n,i,k} \perp \phi_i(q_k) \geq 0, \hspace{1em}
            \norm{\lambda_{t,i,k}}_2 \leq \mu\, \lambda_{n,i,k}.
        \label{eq:contact-conditions}
        \end{equation}
        Here, $M(q) \succ 0$ is the generalized mass/inertia matrix, $b(q,v)$
        collects Coriolis, centrifugal, and gravity terms, $Bu$ is generalized
        actuation with $u \in \mathbb{R}^{n_u}$, $B \in \mathbb{R}^{n_v \times
        n_u}$. 
        Each contact impulse $\bar \lambda_{i,k} = [\lambda_{n,i,k};\lambda_{t,i,k}]
        \in \R^{1+n_t}$ acts through the contact Jacobian
        \begin{equation}
            J_{c,i}(q) = \begin{bmatrix} J_{n,i}(q) \\ J_{t,i}(q) \end{bmatrix} \in \R^{(1+n_t)\times n_v},
            \hspace{0.1em}
            J_{n,i}(q) = \frac{\partial \phi_i}{\partial q} \in \R^{1\times n_v},
            \label{eq:contact-jacobian}
        \end{equation}
        whose top row $J_{n,i}(q)$ is the gradient of the gap function
        $\phi_i(q)$ and $J_{t,i}(q) \in \R^{n_t \times n_v}$ maps $\dot q$ to the relative sliding velocity along each tangent direction. 

        \begin{figure*}[ht!]
            \centering
            \includegraphics[width=\textwidth]{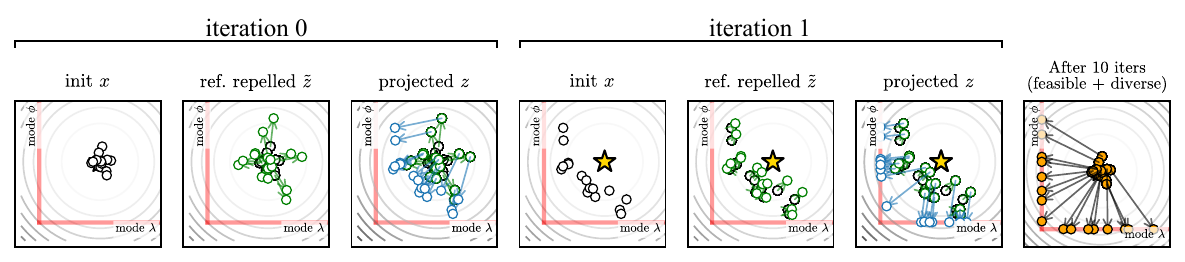}
            \caption{\textbf{Overview of projected variable-splitting construction} on a motivating complementarity
            problem with constraints $0 \;\leq\; \lambda \;\perp\; \phi \;\geq\; 0$, and objective $f = \frac{1}{2}\norm{[\lambda, \phi]^\top - [\lambda^\star, \phi^\star]^\top}_2^2$.
            The initial particles $x$ are first split into duplicate $z$ variables. 
            The Stein repulsion is measured according to $z$ to produce $\tilde z$ as a target reference which is projected onto the feasible set $C$. 
            The $x$-update then tracks the objective while being pulled toward the feasible set $C$ through the split variable.
            }
            \label{fig:complementarity-step-by-step}
        \end{figure*}

    \subsection{Contact-Implicit Trajectory Optimization}
    \label{sec:contact-implicit}

        In practice, \eqref{eq:manipulator}, \eqref{eq:contact-conditions} are specified as equality constraints and further discretized following Stewart and Trinkle~\cite{stewart1996implicit}, so that one step of the rigid-body dynamics together with the discretized form of \eqref{eq:contact-conditions} forms a linear complementarity problem (LCP) in the next velocity and contact impulses.
        Stacking this LCP over a trajectory $x = \{x_0,\dots,x_T\}$ of states $x_k = (q_k,v_k,\{\lambda_{i,k}\},u_k)$ with running cost $\ell$ and terminal cost $\ell_f$ gives the contact-implicit trajectory optimization problem
        \begin{equation}
            \min_x\; \sum_{k=0}^{T-1}\ell(x_k) + \ell_f(x_T) \;\; \text{s.t.} \;\;
            \text{LCP}, \, g(x_k)\leq 0, \, h(x_k) = 0 \,\, \forall k
            \label{eq:contact-implicit}
        \end{equation}
        with boundary conditions $x_0 = x_{init}$, $x_T \in \mathcal{X}_{goal}$, where $g,h$ are additional inequality/equality constraints on the state and control.
        The contact forces are thus lifted as decision variables subject to the LCP~\cite{doi:10.1177/0278364913506757, doi:10.1177/0278364919849235}, letting the optimization reason jointly about contact forces and contact mode rather than relying on pre-specified contact sequence.
        Prior work has shown that ADMM is particularly well-suited for solving contact-implicit trajectory optimization problems, as it can handle the non-smooth LCP constraints~\cite{doi:10.1177/0278364913506757,aydinoglu2024consensus,aydinoglu2021realtime}.

    \subsection{Stein Variational Gradient Descent}

        Given an unnormalized target density $p(x) \propto \exp(-f(x))$ and a set of particles $\{x_i\}_{i=1}^N$ where $x_i \in \R^d$, Stein Variational Gradient Descent (SVGD) moves every particle
        along the direction
        \begin{equation}
        \phi^\ast(x) = \mathbb{E}_{x' \sim q}\big[\, k(x', x)\, \nabla_{x'} \log p(x') \;+\; \nabla_{x'} k(x', x) \,\big],
        \label{eq:svgd-population}
        \end{equation}
        which is the direction, within the unit ball of a reproducing kernel Hilbert space, that most decreases the KL divergence between $q$ (the current empirical particle distribution) and $p$ (the target distribution). 
        One can approximate the direction $\phi^\star$ given a finite set of particles via the empirical average set of particles $\{x_i\}_{i=1}^N$ as
        \begin{multline}
        \phi(x_i) = \frac{1}{N}\sum_{j=1}^N \Big[\, k(x_i, x_j)\, s_j
        +\varepsilon \nabla_{x_j} k(x_i, x_j) \,\Big]
        \label{eq:svgd-empirical}
        \end{multline}
        where $s_j := -\nabla_{x_j} f(x_j)$ is the score and $\epsilon > 0$ is a scaling term.
        The first term drives particles along the score $s_j$ and the second is a repulsive term that pushes $x_i$ away from nearby $x_j$ under the kernel function $k : \R^d \times \R^d \to \R$.
        The particles are then updated \emph{deterministically} with a step size $\gamma$ along the direction as $x_i^\star \leftarrow x_i + \gamma \, \phi(x_i) \, \forall i$ until convergence.
        Note that here, the particles do not satisfy any constraints and  $\phi$ is not aware of any constraints that appear in the score.

    \subsection{Consensus ADMM}
    \label{sec:admm-background}

        Let us consider a generic constrained optimization problem of the form
        \begin{equation}
        \min_x f(x) \quad \text{s.t.} \quad v(x) \in C,
        \label{eq:constrained-problem}
        \end{equation}
        where $f : \R^d \to \R$ is a smooth objective, $v : \R^d \to \R^m$ is a smooth residual function\footnote{This function can be viewed as the constraint equations or can simply be the identity map where $C$ contains the feasible set of $x$.}, and $C \subseteq \R^m$ is a closed convex set.
        The Alternating Direction Method of Multipliers (ADMM) approach to solving~\eqref{eq:constrained-problem} produces a duplicate new variable $z \in \R^m$ to replace the output of $v$ and enforces equality between them as a constraint~\cite{boyd2011distributed}.
        To construct the ADMM formulation, we rewrite \eqref{eq:constrained-problem} as
        \begin{equation}
        \min_{x,z}\; f(x) + \iota_C(z) \quad \text{s.t.} \quad v(x) = z,
        \label{eq:admm-split}
        \end{equation}
        where $\iota_C(v) = 0, \; v \in C$ and $+\infty$ otherwise, is an indicator function that indicates when the constraint is outside the feasible set\footnote{In practice, $\iota_C(v)$  is replaced with an equivalent smooth approximation~\cite{5533136}.}.
        The ADMM algorithm then solves \eqref{eq:admm-split} by forming the augmented Lagrangian
        \begin{equation}
        \mathcal{L}_\rho(x,z,u) = f(x) + \iota_C(z) + \frac{\rho}{2}\norm{v(x) - z + u}_2^2 
        \label{eq:admm-scaled-lagrangian}
        \end{equation}
        where $u \in \mathbb{R}^m$ is the scaled dual variable and $\rho > 0$ is a penalty parameter.
        Solving $x, z$ jointly at this stage reduces to the method of multipliers \cite{bertsekas1999nonlinear}. 
        Instead, ADMM alternates between minimizing $\mathcal{L}_\rho$ over $x$ and $z$ separately, with a dual ascent step on the scaled dual variable $u$.
        More formally, given an initial guess $x^{(0)}, z^{(0)}, u^{(0)}$, the ADMM updates are given as 
        \begin{equation}
            \begin{split}
                x^{(k+1)} \leftarrow \argmin_x \mathcal{L}_\rho(x, z^{(k)}, u^{(k)}), \quad \\
                z^{(k+1)} \leftarrow \argmin_z \mathcal{L}_\rho(x^{(k+1)}, z, u^{(k)}), \quad \\ 
                u^{(k+1)} \leftarrow u^{(k)} + v(x^{(k+1)}) - z^{(k+1)}.            
            \end{split}
            \label{eq:admm-steps}
        \end{equation}
        Note that the $x$-update is unconstrained, while the $z$-update is constrained to $C$ through the indicator function $\iota_C$ which can be expressed as the projection onto $C$
        \begin{equation}
            \begin{split}
                z^\star \leftarrow \argmin_z \Big[\, \iota_C(z) + \tfrac{\rho}{2}\norm{v(x)-z+u}_2^2 \,\Big] \\ 
                    = \proj_C\big(v(x) + u\big),
            \end{split}
        \label{eq:admm-z-general}
        \end{equation}
        which maintains constraint feasibility and converges when a consensus is reached, i.e., when $v(x) = z$.
        Additionally, assuming $f$ is smooth and twice-differentiable, the $x$-update is unconstrained and can be solved with a quasi-Newton method such as L-BFGS or any other suitable optimization algorithm.
        In the following section, we derive the SVGD repulsion term within the $z$-update of ADMM to maintain feasibility and encourage diversity in the particle population.

\section{Stein Projected ADMM}
\label{sec:method}

    Rather than applying the repulsion term directly to the primal variable $x_i$ as in~\eqref{eq:svgd-empirical}, we instead route the repulsion term through the projection of the split variable $z_i$.
    The intuition behind this approach is that the Stein variational gradient is best used as a reference direction within the feasible set of $z_i$, smoothly pulling $x_i$ toward $z_i$ via the dual variable $u$ and ensuring repulsion in the constraint manifold.

    \subsection{Stein Projected Variable Splitting}
    \label{sec:stein-split}

       Given a set of particles $x_i \in \mathcal{X}$, $i = 1,\dots,N$, we instantiate~\eqref{eq:admm-split}--\eqref{eq:admm-z-general} for each particle (and create copies of $z_i$ and $u_i$), which gives the scaled augmented Lagrangian
        \begin{equation}
            \mathcal{L}_\rho(x_i, z_i, u_i) = \mathcal{J}(x_i) + \iota_C(z_i) + \frac{\rho}{2}\norm{v(x_i) - z_i + u_i}_2^2,
            \label{eq:admm-lagrangian}
        \end{equation}
        where we assume a vectorized $x_i$ such that $\| \cdot \|_2$ induces a norm on the space $\mathcal{X}$. 
        Rather than applying the Stein gradient on the primal value $x_i$, we instead compute the Stein gradient with respect to the split variable $z_i$, that is, 
        \begin{align}
            \phi(z_i) & = \frac{1}{N} \sum_{j=1}^N \left[ k(z_i, z_j)s(z_j) + \varepsilon\, \nabla_{z_j}k(z_i, z_j) \right]
            \label{eq:stein-z-grad}
        \end{align}
        where $s(z_j) := -\nabla_{z_j} \mathcal{L}_\rho(x_j^\star, z_j, u_j)$ is the score term applied after the $x$-update.

        Note that blindly applying the Stein gradient to the $z$-update of ADMM would not guarantee that the $z$-update remains feasible, nor would it preserve diversity since the repulsion term can push $z_i$ outside the feasible set $C$, and the projection step would then negate the diversity effect.
        The same is true for applying the Stein gradient to $x_i$, which would not preserve diversity within the feasible set $C$.
        Instead, we inject the Stein gradient~\eqref{eq:stein-z-grad} as a reference target into the $z$-update of ADMM, encouraging diversity within the constraint $C$ while maintaining feasibility.
        More specifically,
        \begin{equation}
        \begin{split}
            z_i^\star \leftarrow &\argmin_z \Big[\, \iota_C(z) + \tfrac{\rho}{2}\norm{\tilde{z}_i + \gamma\,\phi(z_i)-z}_2^2 \,\Big] \\ 
            &= \proj_C\Big(\tilde{z}_i + \gamma\, \phi(z_i)\Big),
        \end{split}
        \label{eq:admm-z-update}
        \end{equation}
        where $\tilde{z}_i = v(x_i^\star) + u_i$, $\gamma > 0$ is a step size that controls the repulsion effect.

        The $x$-update of ADMM remains unchanged, and is given by
        \begin{equation}
            \begin{split}
                x_i^\star \leftarrow & \argmin_{x_i} \mathcal{L}_\rho(x_i, z_i, u_i) \\ 
                & = \argmin_{x_i} \Big[\, \mathcal{J}(x_i) + \frac{\rho}{2}\norm{v(x_i) - z_i + u_i}_2^2 \,\Big].
            \end{split}
            \label{eq:admm-x-update}
        \end{equation}
        which tracks the objective while being pulled toward distinct values $z_i$ within the feasible set $C$.
        The dual variable $u_i$ update remains the same as in standard ADMM,
        \begin{equation}
            u_i^\star \leftarrow u_i + v(x_i^\star) - z_i^\star
        \end{equation}
        and ensures that the $x_i$ and $z_i$ variables reach consensus as the algorithm converges.
        The full algorithm is summarized in Algorithm~\ref{alg:stein-admm} and illustrated in Figure~\ref{fig:complementarity-step-by-step}.

    \begin{algorithm}[h]
    \SetAlgoLined
    \KwIn{initial particles $\{x_i^{(0)}\}_{i=1}^N$, constraint $v(\cdot) \in C$, objective $\mathcal{J}$, kernel $k$, repulsion step size $\gamma$, tolerance $\tau_{1,2}$, max iterations $K$.}
    Initialize $z_i \leftarrow \proj_C\big(v(x_i^{(0)})\big)$, $u_i \leftarrow 0$, $\rho \leftarrow \rho_0$ \\ 
    \For{$k = 0, 1, 2, \dots, K$}{
        \For{$i = 1, 2, \dots, N$ in parallel}{
            \tcp{$x$-update}
            $x_i \leftarrow \argmin \mathcal{L}_\rho(x_i, z_i, u_i)$ \\
            % \tcp{compute Stein gradient on $z$}
            % $\phi(z_i) \leftarrow Eq.~\eqref{eq:stein-z-grad}$ \;
            \tcp{$z$-update}
            $\tilde z_i \leftarrow v(x_i) + u_i$\;
            $z_i \leftarrow \proj_C\big(\tilde z_i + \gamma\, \phi(z_i)\big)$ \\ 
            \tcp{dual update}
            $u_i \leftarrow u_i + v(x_i) - z_i$ \\
        }
    $\forall i$ \lIf{$\norm{\nabla_{x_i} \mathcal{L}_\rho}_\infty \leq \tau_1$ and $\norm{\dist_C\big(v(x_i)\big)}_\infty \leq \tau_2$}{\textbf{break}}
    }
    \KwOut{$\{x_i\}$ and $\{z_i\}$}
    \caption{Stein Projected ADMM}
    \label{alg:stein-admm}
    \end{algorithm}

\section{Results} 
\label{sec:results}

    In this section we test the effectiveness of the proposed Stein Projected ADMM on a number of motivational problems and contact-rich manipulation tasks. All findings are summarized by bold paragraph headers.

    \subsection{Motivating Examples: Annulus and Complementarity}
    \label{sec:annulus}

       \begin{figure}[h]
        \centering
        \includegraphics[width=\columnwidth]{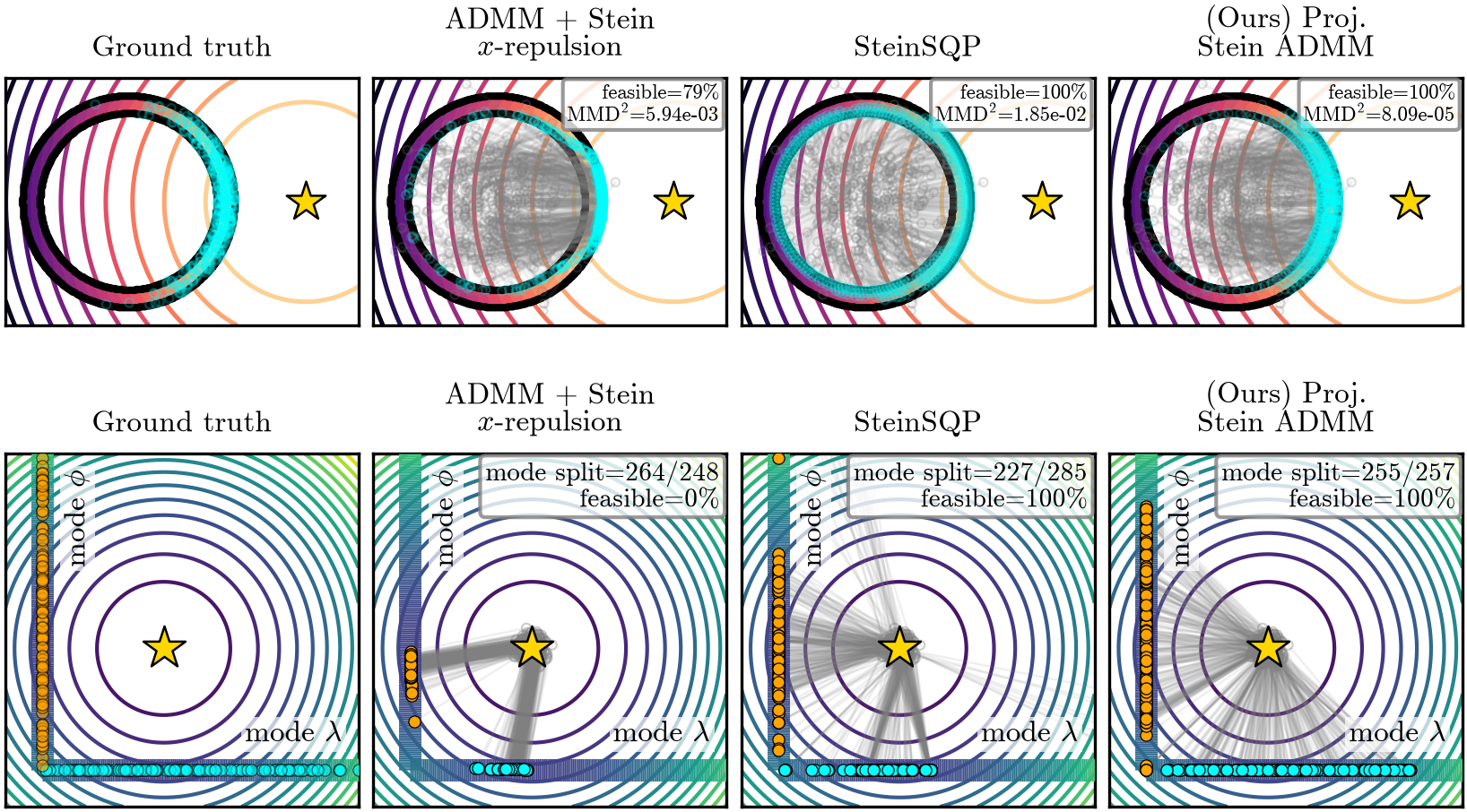}
        \caption{Illustrative example of $N=512$ particles, on the annulus problem (top) and the complementarity problem (bottom). At large samples, the proposed approach evenly spreads particles within the feasible set within tolerance. Baseline methods have comparable performance at low-particle numbers but overemphasize certain regions with larger particle sizes.}
        \label{fig:toy-examples}
        \end{figure}

        We test the projected variable-splitting construction of Section~\ref{sec:stein-split} on two motivating examples: a 2-D annulus non-convex feasible band with a quadratic objective pulling particles toward a point outside the band~\cite{li2026globalizedconstrainedsteinvariational}; and a complementarity-constrained problem defined with the complementarity $0 \leq \lambda \perp \phi \geq 0$, with a quadratic objective given by $f = \frac{1}{2}\norm{[\lambda, \phi]^\top - [\lambda^\star, \phi^\star]^\top}_2^2$ that evenly drags particles towards modes $[\lambda^\star, 0]^\top$ or $[0, \phi^\star]^\top$. In both cases, the failure mode is that the repulsion term gathers particles along the boundary of the feasible set (closest to the local optima), rather than spreading evenly across the feasible set.

        \begin{table}[h]
        \centering
        \caption{Annulus problem, median over $10$ seeds.}
        \label{tab:annulus}
        \footnotesize
        \begin{tabular}{lcccc}
        \toprule
        Method & $N$ & Iters. & Feasible & MMD$^2$ \\
        \midrule
        Stein ADMM $x$-repulsion & 88 & 400 & 82.4\%  & $1.64\times10^{-3}$ \\
        Stein ADMM $z$-repulsion & 88 & 495 & 77.3\%  & $3.77\times10^{-4}$ \\
        SteinSQP~\cite{li2026globalizedconstrainedsteinvariational} & 88 & 238 & 100\% & $2.52\times10^{-3}$ \\
        \midrule
        Stein Projected ADMM & 88 & 235 & 100.0\% & $1.29\times10^{-4}$ \\
        \bottomrule
        \end{tabular}
        \end{table}

        \vspace{0.3em}
        \noindent
        \textbf{Even particle spread.} As illustrated in Fig.~\ref{fig:complementarity-step-by-step}, the proposed projected variable-splitting construction decouples the repulsion term from the primal $x$-update, allowing the repulsion to be measured in the split variable $z$ and then projected onto the feasible set $C$.
        This naturally encourages diversity within the feasible set and takes advantage of the dual variable within ADMM to pull the solution toward feasibility without compromising the repulsion effect.
        In contrast, we find that applying the Stein gradient to the $x$-update or $z$-update as a naive baseline places too much emphasis on the objective, clumping particles along the boundary of the feasible set and leaving a large fraction of particles infeasible (Fig.~\ref{fig:toy-examples}).

    \begin{figure}[h]
    \centering
    \includegraphics[width=\columnwidth]{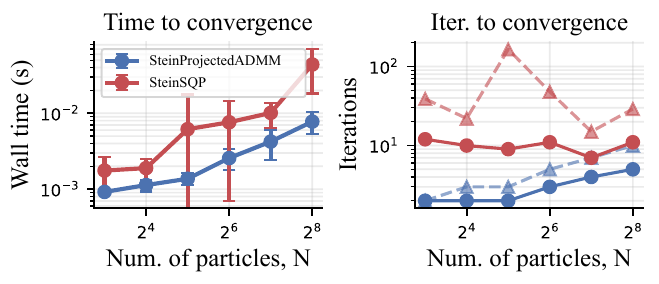}
    \caption{Numerical comparison on complementarity problem. (Left) Wall-clock time to first reach constraint $\text{tol}=10^{-4}$ vs.\ particle
    count $N$ (over $9$ seeds). (Right) iterations to
    convergence for Stein Projected ADMM (ours)
    and SteinSQP~\cite{li2026globalizedconstrainedsteinvariational}. Experiments were conducted on an Apple M4 Pro chip.}
    \label{fig:complementarity-scaling}
    \end{figure}

        \vspace{0.3em}
        \noindent
        \textbf{Favorable scaling at larger particle numbers.}
        Interestingly, SteinSQP~\cite{li2026globalizedconstrainedsteinvariational} does perform similarly to the proposed approach at low particle numbers as shown in Table~\ref{tab:annulus}, but overemphasizes certain regions of the feasible set at larger particle numbers (see Fig.~\ref{fig:toy-examples}).
        This is further emphasized in the complementarity problem, where SteinSQP struggles to spread particles evenly across the two modes and required twice the computation time (see Table~\ref{tab:complementarity} and Fig.~\ref{fig:complementarity-scaling}).
        When comparing the wall-clock time to reach tolerance, the proposed approach scales competitively with particle number and is approximately $2\times$ faster than SteinSQP with less variability. 
        This can be attributed to the fact that SteinSQP forms a constrained optimization problem that requires a line search which can induce variability in the number of iterations to convergence. 
        Our approach, on the other hand, leverages the ADMM variable splitting to smooth out the pull between the objective, constraint, and Stein repulsive term, allowing for more consistent number of iterations to convergence as shown in Fig.~\ref{fig:complementarity-scaling}.

    \begin{table*}[h]
    \centering
    \vspace{0.5em}
    \caption{Complementarity problem ($0 \le \phi \perp \lambda \ge 0$), median
    over $10$ seeds.}
    \label{tab:complementarity}
    \footnotesize
    \resizebox{\textwidth}{!}{%
    \begin{tabular}{lccccccc}
    \toprule
    Method & Feasible & MMD$^2$ & Convergence (Iters.). & Term. Viol. & ms/iter & Mode $\phi$ & Mode $\lambda$ \\
    \midrule
    Stein ADMM $x$-repulsion & 0.0\% & $1.97\times10^{-2}$ & 200 (never) & $6.74\times10^{-3}$ & $0.51\pm0.03$ & $36.0\pm3.7$ & $30.0\pm3.7$ \\
    Stein ADMM $z$-repulsion & 88.6\% & $6.72\times10^{-4}$ & 36 & $9.39\times10^{-5}$ & $0.60\pm0.01$ & $32.5\pm1.1$ & $33.5\pm1.1$ \\
    SteinSQP~\cite{li2026globalizedconstrainedsteinvariational} & 100.0\% & $3.21\times10^{-3}$ & 6 & $2.99\times10^{-5}$ & $1.00\pm0.04$ & $30.0\pm3.8$ & $36.0\pm3.8$ \\
    \midrule
    Stein Projected ADMM & 100.0\% & $3.41\times10^{-3}$ & 3 & $6.03\times10^{-5}$ & $0.54\pm0.01$ & $33.0\pm0.7$ & $33.0\pm0.7$ \\
    \bottomrule
    \end{tabular}%
    }
    \end{table*}

    \begin{figure}[h]
    \centering
    \includegraphics[width=\columnwidth]{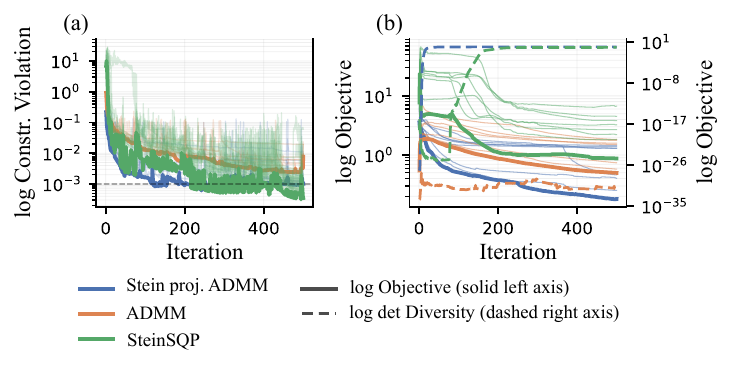}
    \caption{Convergence curves of the box pivoting problem. (a) Maximum constraint violation for each particle solution. (b) Diversity measured by~\eqref{eq:diversity} and objective value per trajectory. Faded colors are the individual constraint curves, solid colors are the mean of the particle set, mean objective (left axis, solid) and
    log-determinant diversity (right axis, dashed). }
    \label{fig:box-pivoting-convergence}
    \vspace{-0.5em}
    \end{figure}

  \subsection{Box Pivoting with Two Fingers}
  \label{sec:box-pivoting}

    We next evaluate the proposed Stein Projected ADMM on a contact-rich manipulation task. 
    Here, we consider the box pivoting with two fingers task as described in~\cite{aydinoglu2021realtime}.
    The goal is to pivot a box from an initial pose to a target pose using two fingers, while maintaining contact and respecting friction constraints.
    We compare the proposed Stein Projected ADMM against an ADMM baseline with no repulsion and SteinSQP~\cite{li2026globalizedconstrainedsteinvariational} on the same randomized initialization and problem setup.

    \begin{figure}[h]
    \centering
    \includegraphics[width=\columnwidth]{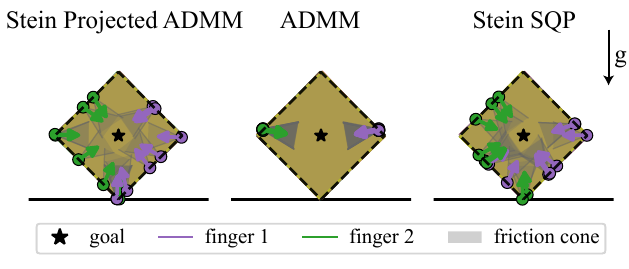}
    \caption{Final overlaid configurations and finger placements in the box pivoting problem for Stein Projected ADMM (ours), the
    repulsion-off ADMM baseline, and SteinSQP~\cite{li2026globalizedconstrainedsteinvariational}.}
    \label{fig:box-pivoting-final-config}
    \end{figure}

    \vspace{0.3em}
    \noindent
    \textbf{Earlier ensemble convergence at better optima.}
    We first evaluate the convergence of the three methods in terms of constraint violation, mean objective value, and diversity over iteration count as shown in Fig.~\ref{fig:box-pivoting-convergence}.
    The proposed Stein Projected ADMM achieves competitive constraint violation and higher diversity (measured using the log-determinant of the ensemble covariance, see Appendix, Equation~\ref{eq:diversity}) compared to the ADMM baseline with no repulsion and competing with SteinSQP.
    The main advantage of the proposed method is that it achieves a lower objective value across the particle ensemble while maintaining feasibility and diversity, which is critical for contact-rich manipulation tasks where multiple feasible solutions exist.
    An illustration of the final box configurations and terminal contact forces for each method is shown in Fig.~\ref{fig:box-pivoting-final-config}.

  \subsection{Non-prehensile Pushing}
  \label{sec:cube-push-kernel-solver}
    
    This next experiment evaluates the proposed Stein Projected ADMM on a non-prehensile pushing task where the goal is to evaluate the effect of different SVGD kernels on the diversity of the resulting cube-pushing trajectories.
    We compare three different kernels: RBF, Cauchy, and Laplace, as well as SteinSQP~\cite{li2026globalizedconstrainedsteinvariational} with contact as an explicit variable that is being optimized. \footnote{This differs from the original implementation of the same experiment in~\cite{li2026globalizedconstrainedsteinvariational} where contact is not an explicit variable that is optimized.}

    % --- single-column version ---
    \begin{figure}[h]
    \centering
    \includegraphics[width=\columnwidth]{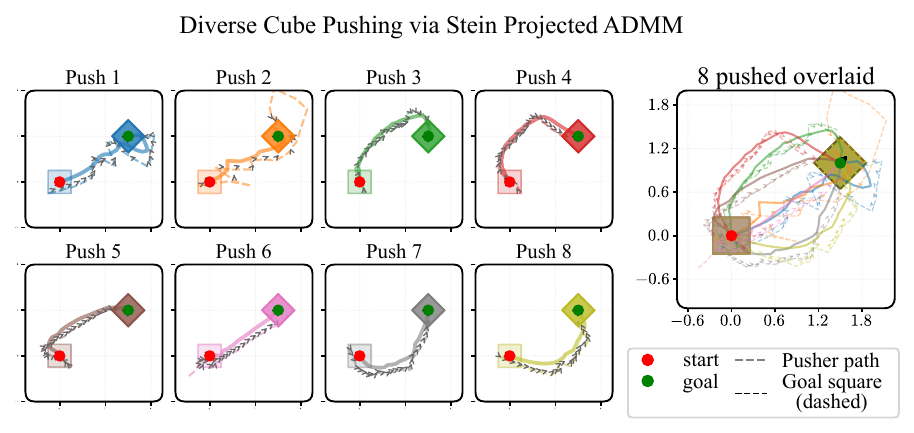}
    \caption{Diverse cube-pushing trajectories generated from Stein Projected ADMM. Each trajectory particle reaches the same goal pose via different contact strategies. (Left) Rendered particle trajectories with contact forces and points. (Right) Overlaid trajectory distribution.}
    \label{fig:cube-push-1col}
    \end{figure}

\begin{figure}[t]
    \centering
    \includegraphics[width=\columnwidth]{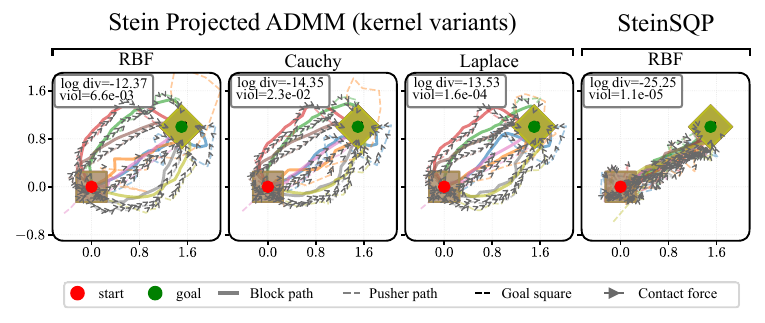}
    \caption{Diverse cube-pushing trajectories on the same initialization
    under three SVGD kernels (RBF, Cauchy, Laplace) and
    \cite{li2026globalizedconstrainedsteinvariational}'s SteinSQP with contact as an explicit variable (differing from the original implementation).}
    \label{fig:cube-push-kernel-solver}
    \end{figure}

    \vspace{0.3em}
    \noindent
    \textbf{Diversity under different kernels.}
    We evaluate the diversity of the resulting cube-pushing trajectories under each kernel by measuring the log-determinant of the ensemble covariance of the final cube poses across all particles.
    We first illustrate the resulting trajectories under the RBF kernel in Fig.~\ref{fig:cube-push-1col}, where we can see that the particles spread out across different contact strategies to reach the same goal pose.
    This includes pushing the cube from different sides, using different contact points, and applying different forces.
    We then compare the diversity of the resulting trajectories under the three different kernels and SteinSQP~\cite{li2026globalizedconstrainedsteinvariational} under the RBF kernel in Fig.~\ref{fig:cube-push-kernel-solver}.
    As the problem is non-convex, we observe that the RBF kernel subtly produces the most diverse set of trajectories, while the Cauchy and Laplace kernels observe little difference. 
    The main variability is in the constraint satisfaction which differs across the three kernels as shown in Table~\ref{tab:cube-push-kernel-solver} and can be adjusted by tuning the penalty parameter $\rho$.
    
    When comparing against SteinSQP~\cite{li2026globalizedconstrainedsteinvariational}, we observe that the proposed Stein Projected ADMM achieves better diversity while maintaining feasibility, whereas SteinSQP~\cite{li2026globalizedconstrainedsteinvariational} achieves lower constraint violation but at the cost of reduced diversity.
    This is likely due to the fact that SteinSQP~\cite{li2026globalizedconstrainedsteinvariational} has been modified to optimize for contact as an explicit variable which increases the overall dimensionality and creates an imbalance between constraint satisfaction and encouraging solution diversity.
    The variable splitting construction of the proposed Stein Projected ADMM allows for a more balanced trade-off between constraint satisfaction and diversity, as diversity is injected more directly into the constraint space.

    \begin{table}[t]
    \centering
    \caption{Kernel/solver comparison on the cube-pushing trajectory
    problem ($N{=}8$, 300 iterations).}
    \label{tab:cube-push-kernel-solver}
    \footnotesize
    \resizebox{\columnwidth}{!}{%
    \begin{tabular}{lcccc}
    \toprule
    Method & Max $|eq|$ & Max ineq & Diversity & Mean obj. \\
    \midrule
    RBF    & $6.62\times10^{-3}$ & $3.22\times10^{-3}$ & $-12.37$ & $0.037$ \\
    Cauchy   & $2.33\times10^{-2}$ & $7.67\times10^{-4}$ & $-14.35$ & $0.030$ \\
    Laplace & $1.12\times10^{-4}$ & $1.58\times10^{-4}$ & $-13.53$ & $0.036$ \\
    SteinSQP (RBF)~\cite{li2026globalizedconstrainedsteinvariational} & $1.50\times10^{-7}$ & $1.10\times10^{-5}$ & $-25.25$ & $0.232$ \\
    \bottomrule
    \end{tabular}%
    }
    \end{table}

  \subsection{Quasi-static Grasping and Humanoid Hand-over}
  \label{sec:quasi-static-grasping-hand-over}

    Finally, we illustrate the proposed Stein Projected ADMM on a quasi-static grasping and humanoid hand-over task.
    Both these tasks are object-centric in that the contact points and forces are solved for in order to achieve quasi-static equilibrium for a target object pose while respecting friction and contact constraints.
    Differential kinematic constraints, manipulability, and collision are then taken into account in the optimization to ensure feasibility within lower-level coordination that is handled by the kinematics solver, similar to~\cite{Zakka_Mink_Python_inverse_2026}.

    \begin{figure*}[t]
        \centering
        \vspace{0.5em}
        \includegraphics[width=\linewidth,trim=0 0 0 1.3em,clip]{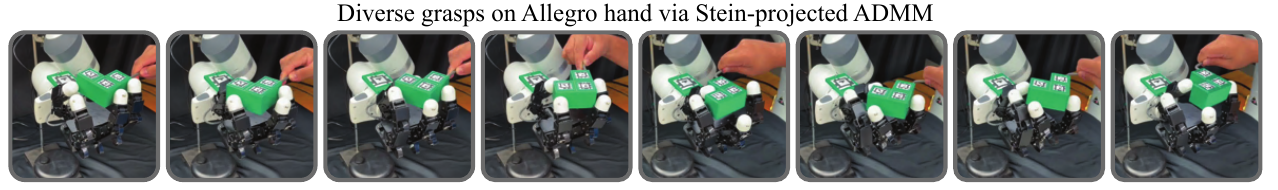}
        \caption{Diverse grasps on Allegro hand hardware solved with Stein-projected ADMM. We evaluate four-fingered quasi-static grasps of a non-convex object that runs in real-time (we refer the reader to the multimedia material). The executed grasp is chosen via an optimistic (best cost) heuristic from the solved candidate grasps.}
        \label{fig:quasi-static-grasping-hand-over}
    \end{figure*}

    \vspace{0.3em}
    \noindent
    \textbf{Diverse grasps and whole-body coordination:}
    We illustrate the resulting diverse feasible grasps from a run of Stein Projected ADMM in Fig.~\ref{fig:quasi-static-grasping-hand-over} with an Allegro hand with four finger point contacts.
    We can see that the primal solutions spread out across different contact regions to reach the same quasi-static equilibrium, including different finger placements, orientations, and forces.
    This also enables the generation of diverse feasible grasps and manipulation strategies in a variety of scenarios.
    The multi-body kinematics are incorporated into the ADMM formulation by defining the constraint function $v(x)$ or penalties that comprise the kinematic constraints of the multi-body system, and the projection onto the feasible set $C$ can be performed using a suitable differential kinematic solver that respects the joint limits and contact constraints of the system.
    See Fig.~\ref{fig:teaser} for an illustration of the resulting diverse feasible grasps.

\section{Conclusion}
\label{sec:conclusion}

    This paper derives a novel variable splitting approach that integrates Stein variational inference into ADMM.
    The proposed approach applies the Stein gradient as a reference target for the split variable $z_i$ rather than the primal variable $x_i$, decoupling repulsion from the primal update and keeping it within the feasible set.
    We demonstrate the effectiveness of the proposed method across two motivating problems and four contact-rich manipulation tasks (e.g., box pivoting, non-prehensile pushing, quasi-static grasping, and humanoid hand-over) and show that this construction achieves better diversity than naive repulsion and existing methods without sacrificing feasibility.
    Last, we show that the proposed method scales favorably with particle number and is competitive with existing methods in terms of wall-clock time to convergence and the diversity of solutions.

    % Three directions follow: an adaptive rule for kernel bandwidth and penalty
    % scheduling would remove the manual tuning that currently governs this
    % trade-off; the hand-specified projection used for grasping and hand-over
    % could be constructed automatically to generalize to other multi-body
    % systems; and since every result here is obtained in simulation, evaluating
    % the method within a real-time or hardware-in-the-loop pipeline is a
    % natural next step given its comparatively simple, favorably-scaling
    % per-particle update.

\appendix

\noindent\textbf{Polyhedral friction-cone approximation.}\label{app:friction}
The friction cone is approximated via a polyhedron using fixed basis vector $D_k$ spanning the tangent plane.
The idea is to compose the friction force as a non-negative combination of a finite set of tangent directions rather than an arbitrary vector inside the cone,
$
J_{t,k}^\top \lambda_{t,k} = D_k \beta_k, \beta_k \geq 0,
\label{eq:friction-basis}
$
where $\beta_k$ are the non-negative coefficients, and the subscript $k$ indicates the time step.
A slip-rate variable $\zeta_k$ is introduced to enforce the maximum dissipation principle, which ensures friction opposes relative sliding at the boundary of its cone.
This forms the following complementarity conditions
\begin{equation*}
0 \leq \beta_k \perp \zeta_k \mathbf{1} + D_k^\top v_{k+1} \geq 0, \quad
0 \leq \zeta_k \perp \mu\lambda_{n,k} - \mathbf{1}^\top\beta_k \geq 0.
\label{eq:st-friction}
\end{equation*}
where $\mathbf{1}$ is a vector of ones, and $\mu$ is the friction coefficient.
The complementarity conditions are defined in $(\beta_k, \zeta_k)$.

\noindent\textbf{Measuring diversity.}\label{app:diversity}
Diversity within the particle ensemble is measured by the log-determinant of the kernel Gram matrix given by
\begin{equation} \label{eq:diversity}
    D(\{x_i\}_{i=1}^N) = \log\det\big(K + \varepsilon I\big)
\end{equation}
for $K_{ij} = k(x_i, x_j; h)$ and $\varepsilon>0$ is a small regularization term.
Here, a larger value (approaching zero) of the log-determinant indicates a more diverse set of particles.
The smaller (and more negative) value indicates that the particles are clustered together and not diverse.
The choice of kernel and bandwidth $h$ can affect the sensitivity of this metric, especially in high-dimensional (trajectory) spaces, and is chosen to reflect the scale of the problem.

\noindent\textbf{Annulus problem.}\label{app:annulus-detail}
The objective pulls particles toward a point outside the feasible band,
\begin{equation}
    f(x) = \frac{1}{2\sigma^2}\norm{x-\mu}_2^2, \quad \mu = (5,0)^\top,\ \sigma = 2,
\end{equation}
subject to the non-convex feasible band $C=\{x\in\R^2: r_{in}\le\norm{x}_2\le r_{out}\}$, expressed as two inequalities
\begin{equation}
    g_1(x) = r_{in}^2 - \norm{x}_2^2 \le 0, \quad g_2(x) = \norm{x}_2^2 - r_{out}^2 \le 0,
\end{equation}
with $r_{in}=2.5$, $r_{out}=3$. Particles are initialized $x_i^{(0)}\overset{\text{iid}}{\sim}\mathcal N(0,I)$, $N=88$, with penalty $\rho=100$, fixed RBF bandwidth $h=0.01$, tolerance $10^{-4}$. All methods use repulsion $\gamma=1.0$ and baselines set to default configuration.

\vspace{0.3em}
\noindent\textbf{Complementarity problem.}\label{app:complementarity-detail}
Let $x=(\lambda,\phi)\in\R^2$, the objective is a quadratic centered on the infeasible point $\mu=(\lambda^\star,\phi^\star)=(1,1)$ defined as
$
f(x) = \tfrac12\norm{x-\mu}_2^2,
$
whose minima are projected evenly between $[\lambda^\star,0]^\top$, $[0,\phi^\star]^\top$ by the exact complementarity constraint
\begin{equation}
    \lambda\phi = 0 \ (\text{equality}), \qquad \lambda\ge0,\ \phi\ge0 \ (\text{inequality}).
\end{equation}
Particles start collapsed at $x_i^{(0)} \sim \mathcal N(\mu, 0.05 I)$ with a fixed bandwidth $h=0.02$, $\rho=100$, and solver tolerance $10^{-4}$. All methods use the same $\gamma=0.1$. 

\vspace{0.3em}
\noindent\textbf{Box pivoting with two fingers.}\label{app:box-pivot-detail}
Writing $q_t=(x_t,y_t,\alpha_t)$ for the box pose and $p_{1,t},p_{2,t}$ for the two finger contact points, the objective is defined as 
\begin{equation}
% \begin{split}
%     \mathcal{J} = \frac{1}{2(T-1)} \sum_{t=0}^{T-2}  & w_c \norm{p_{1,t+1}-p_{1,t}}_2^2
%     + w_c \norm{p_{2,t+1}-p_{2,t}}_2^2 \\ 
%     & + w_b \norm{q_{t+1}-q_t}_2^2 ,
% \end{split}
\begin{split}
    \mathcal{J} = \frac{1}{2(T-1)} \sum_{t=0}^{T-2} \sum_{n=1}^2  & w_c \norm{p_{n,t+1}-p_{n,t}}_2^2 \\ 
    & + w_b \norm{q_{t+1}-q_t}_2^2 ,
\end{split}
\end{equation}
where $w_c=10$, $w_b=1$. The terminal orientation is handled as a two-sided inequality band $|\alpha_T-\pi/4|\le1^\circ$, letting particles settle at different final tilts. Inequality constraints are given by \eqref{eq:contact-conditions} that enforce complementarity and friction cone on the contact force variables for each finger and ground contact. Friction coefficients are given as $\mu_{\text{finger}}=0.1$, $\mu_{\text{ground}}=1.0$, with $T=25$, $\Delta t=0.05$s for all methods. All methods use $\gamma=0.4$ with $\rho=100$ and RBF kernel is evaluated on the terminal contact positions with $h=0.2$.

\vspace{0.3em}
\noindent\textbf{Non-prehensile pushing.}\label{app:cube-push-detail}
A single pusher contacts the box at body-frame point $c_t$ with pose $q_t=(x_{b,t},y_{b,t},\theta_t)$ for the box pose and $q_{b_T}^\star$ for the target. The objective is given as
\begin{equation}
\begin{split}
    J = \frac{1}{2(T-1)} \sum_{t=0}^{T-2} w_c \norm{c_{t+1}-c_t}_2^2
    + w_b \norm{q_{t+1}-q_t}_2^2 
    \\
    + \frac{w_T}{2}\Big(\norm{b_T - q_{b_T}^\star}_2^2 + (\theta_T-\theta_{b_T}^\star)^2\Big),
\end{split}
\end{equation}
$w_c=w_T=10$, $w_b=1$. The pusher follows quasi-static equilibrium (via equality constraints)
% , $b_{t+1}-b_t=\Delta t\,F_t$, $\theta_{t+1}-\theta_t = \Delta t\,\tau_t/I_{\text{eff}}$ with $F_t=\lambda_{n,t}n_t+\lambda_{t,t}\,t_t$, $\tau_t = r_{x,t}F_{y,t}-r_{y,t}F_{x,t}$, $I_{\text{eff}}=0.25$, plus boundary pins $q_0=p_0$, $q_T=p_T$. 
Contact condition set by $\phi=\ell_\infty(c)-0.25$. Contact constraints are given by \eqref{eq:contact-conditions}. Friction given by $\mu=0.5$, the time horizon $T=25$, $\Delta t=0.1$s. All methods set $\gamma=0.2$, $\rho=100$, The kernel study (Table~\ref{tab:cube-push-kernel-solver}) compares the median-bandwidth RBF kernel against Cauchy $k(x,x';h)=\big(1+\norm{x-x'}^2/h\big)^{-1}$ and Laplace $k(x,x';h)=\exp\!\big(-\norm{x-x'}_1/h\big)$ kernels, with SteinSQP optimizing over $(q,c,\lambda_n,\lambda_t)$.

\vspace{0.3em}
\noindent\textbf{Quasi-static grasping.}\label{app:grasp-detail}
Here, the objective is set to zero $\mathcal J\equiv0$ and diversity is generated entirely by Stein repulsion inside the force-closure manifold. The point-finger contacts at positions $c_i$ with normal/tangential forces $\lambda_{n,i},\lambda_{t,i}$ satisfy the equality constraints and planar force balance, torque balance about the object centroid, and each contact lying on the object surface.
% \begin{equation}
%     \sum_{i=1}^3\big({-}\lambda_{n,i}\,n_i+\lambda_{t,i}\,t_i\big) = 0, \quad
%     \sum_{i=1}^3 r_i\!\times\! F_i = 0, \quad \phi(c_i)=0,
% \end{equation}
% $r_i=c_i-\text{centroid}$. 
Friction is given by $\mu=0.5$, and all methods use $\gamma=0.1$ with RBF kernel.

\vspace{0.3em}
\noindent\textbf{Humanoid hand-over.}\label{app:handover-detail}  Involves two humanoid agents transferring an object between them. The objective is again set to zero, $\mathcal J\equiv0$, and diversity is generated by Stein repulsion within the feasible set defined by the equality and inequality constraints.
There are $4$ hand contacts (two per humanoid) with forces $F_h = -\lambda_{n,h}\hat n_h+\lambda_{t1,h}\hat t_{1,h}+\lambda_{t2,h}\hat t_{2,h}$, which need to satisfy quasi-static equilibrium of the box about its geometric center, 
% with 
% the fixed midpoint $X_H$ between the two torsos,
% \begin{equation*}
%     \sum_{h=1}^4 F_h + (0,0,{-}mg) = 0, \qquad \sum_{h=1}^4 (c_h-X_H)\times F_h = 0
% \end{equation*}
and must satisfy complementarity, friction, and an arm-reach bound $\norm{c_h-\text{shoulder}_h}_\infty\le r_{\text{reach}}$ constraints.
The cube mass is given by $m=3$kg with a half-width of $0.18$m and friction coefficient $\mu=0.5$.
All methods use $\gamma=0.4$ for the repulsion step and RBF kernel.

{\footnotesize
\renewcommand{\baselinestretch}{0.92}
\selectfont
\bibliographystyle{IEEEtran}
\bibliography{main}
}

\end{document}